\documentstyle[qtree,aclap]{article}

\title{\vspace{-0in}Three Generative, Lexicalised Models for Statistical Parsing}
\author{Michael Collins\thanks{This research was supported by ARPA 
Grant N6600194-C6043.}
\\ Dept. of Computer and Information Science \\
University of Pennsylvania \\ Philadelphia, PA, 19104, U.S.A. \\
{\tt mcollins@gradient.cis.upenn.edu}}

\begin{document}
\setlength{\topmargin}{-0.5in}
\setlength{\textheight}{9.5in}
\setlength{\headheight}{-0.5in}
\input{psfig.tex}
\maketitle
\vspace{-0.5in}
\begin{abstract}

In this paper we first propose a new statistical parsing
model, which is a generative model of lexicalised context-free
grammar. We then extend the model to include a probabilistic treatment
of both subcategorisation and wh-movement. Results on Wall Street Journal text
show that the parser performs at 88.1/87.5\% constituent precision/recall,
an average improvement of 2.3\% over \cite{collins}.
%In this paper we first propose a new statistical parsing
%model, which is a generative model of lexicalised context-free
%grammar. We then extend the model to include a probabilistic treatment
%of both subcategorisation and wh-movement. 
%The work is important for two reasons: First, 
%the final results on Wall Street Journal text 
%are 88.1/87.5\% constituent precision/recall,
%an average improvement of 2.3\% over \cite{collins}, which as far as
%we know has the best published results on this task.
%Second, the best performing parsers on WSJ text have, until now,
%ignored information about wh-movement and subcategorisation.
%Most NLP applications will need this information to extract
%predicate-argument structure from parse trees.
\end{abstract}

\bibliographystyle{fullname}

\section{Introduction}

Generative models of syntax have been central in linguistics since
they were introduced in \cite{chomsky}. Each sentence-tree pair $(S,T)$ in a
language has an associated top-down derivation consisting of a
sequence of rule applications of a grammar. These models can be extended to be
statistical by defining probability distributions at points of
non-determinism in the derivations, thereby assigning a probability
${\cal P}(S,T)$ to each $(S,T)$ pair.
Probabilistic context free grammar \cite{pcfg} was an early example
of a statistical grammar. A PCFG can be lexicalised by associating a
head-word with each non-terminal in a parse tree; thus far, 
\cite{magerman,ibm} and \cite{collins}, which both make heavy use of lexical
information, have reported the best statistical parsing performance on
Wall Street Journal text. Neither of these models is generative,
instead they both estimate ${\cal P}(T\:  | \: S)$ directly.

\setlength{\topmargin}{0in}
\setlength{\textheight}{9in}

This paper proposes three new parsing models. {\bf Model 1}
is essentially a generative version of the model described in \cite{collins}.
In {\bf Model 2}, we extend the parser to make the
complement/adjunct distinction by adding probabilities over
subcategorisation frames for head-words. In {\bf Model 3} we give a
probabilistic treatment of wh-movement, which is derived from the
analysis given in Generalized Phrase Structure Grammar \cite{gpsg}.
The work makes two advances over previous models: First, Model
1 performs significantly better than \cite{collins}, and Models 2 and 3
give further improvements --- our final results are 88.1/87.5\% constituent
precision/recall, an average improvement of 2.3\% over \cite{collins}. Second,
the parsers in \cite{collins} and \cite{magerman,ibm} produce trees without
information about wh-movement or subcategorisation.
Most NLP applications will need this information to extract
predicate-argument structure from parse trees.

%In the remainder of this paper we describe the 3 models in
%section~\ref{sec-models}, discuss the practical issues of smoothing
%and parsing in section~\ref{sec-prac}, give results on Wall Street
%Journal text with the parser trained and tested
%on the Penn treebank \cite{marcus} in section~\ref{sec-results}, and finally
%give conclusions in section~\ref{sec-conc}.

In the remainder of this paper we describe the 3 models in 
section~\ref{sec-models}, discuss practical issues in   
section~\ref{sec-prac}, give results in section~\ref{sec-results},
and give conclusions in section~\ref{sec-conc}.

\section{The Three Parsing Models}

\begin{figure*}[tb]
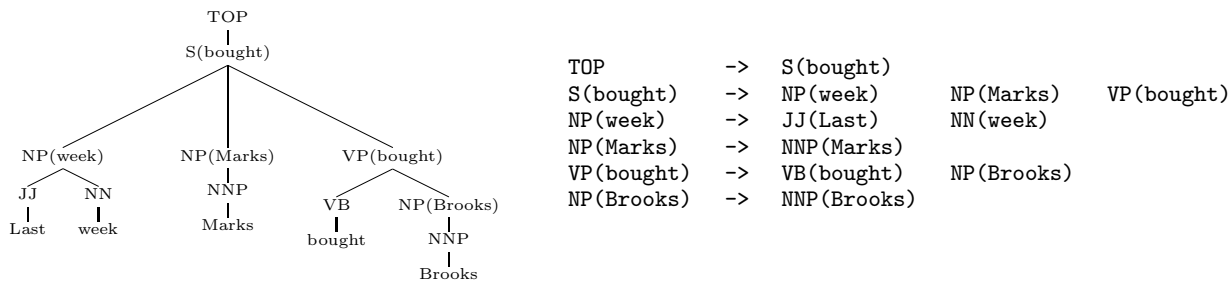

\qtreecenterfalse
\vspace{-0.8in}
\hspace{-0.4in}
\begin{tiny}
\Tree  [.TOP [.S(bought) [.NP(week)  [.JJ Last ]  [.NN week ]  ]
[.NP(Marks)  [.NNP Marks ]  ]
[.VP(bought)  [.VB bought ] [.NP(Brooks)  [.NNP Brooks ]  ] ] ] ]
\end{tiny} 
\begin{small}
\begin{tt} \hspace{0in}
\begin{tabular}{lllll}
&&&&\\
&&&&\\
&&&&\\
&&&&\\
&&&&\\
&&&&\\
&&&&\\
&&&&\\
&&&&\\
TOP          &->   &  S(bought) &   &
\\
S(bought) &->   &NP(week) &NP(Marks) &VP(bought)
\\
NP(week)    &->   &JJ(Last) &NN(week)&
\\
NP(Marks)    &->   &NNP(Marks) &&
\\
VP(bought)   &->   &VB(bought) &  NP(Brooks)&
\\
NP(Brooks) &-> & NNP(Brooks) &&
\end{tabular}
\end{tt}
\end{small}
\caption{A lexicalised parse tree, and a list of the rules it contains.
For brevity we omit the POS tag associated with each word.}
\label{fig-lextree}
\end{figure*}

\label{sec-models}

\subsection{Model 1}

In general, a statistical parsing model defines the conditional
probability,  ${\cal P}(T\:  | \: S)$, for each candidate parse tree $T$ for a
sentence $S$. The parser itself is an algorithm
which searches for the tree, $T_{best}$, that maximises 
${\cal P}(T\:  | \: S)$. A generative model uses the observation that
maximising ${\cal P}(T,S)$
is equivalent to maximising ${\cal P}(T\:  | \: S)$:
\footnote{${\cal P}(S)$ is constant, hence maximising $\frac{{\cal P}(T,S)}{{\cal P}(S)}$
is equivalent to maximising ${\cal P}(T,S)$.}
\begin{eqnarray}
T_{best} &=& \arg \max_T {\cal P}(T\:  | \: S)
= \arg \max_T \frac{{\cal P}(T,S)}{{\cal P}(S)} \nonumber \\
&=& \arg \max_T {\cal P}(T,S)
\end{eqnarray}
${\cal P}(T,S)$ is then estimated by attaching probabilities to a top-down
derivation of the tree.
In a PCFG, for a tree derived by $n$ applications of
context-free re-write
rules $LHS_i \Rightarrow RHS_i$, $1 \leq i \leq n$, 
\begin{equation}
{\cal P}(T,S) = \prod_{i=1..n} {\cal P}(RHS_i\:  | \: LHS_i)
\end{equation}
The re-write rules are either internal to
the tree, where $LHS$ is a non-terminal and $RHS$ is a string of one or more
non-terminals; or lexical, where $LHS$ is a part of speech tag and
$RHS$ is a word.

\setlength{\headheight}{0in}

A PCFG can be lexicalised\footnote{We find lexical heads in Penn
treebank data using rules which are similar to those used by \cite{magerman,ibm}.}
by associating a word $w$ and a
part-of-speech (POS) tag $t$ with each non-terminal $X$ in the tree.
Thus we write
a non-terminal as $X(x)$, where $x=\langle w,t \rangle$, and $X$ is a
constituent label. Each rule
now has the form\footnote{With the exception of the top rule in the tree, 
which has the form {\tt TOP $\rightarrow$ H(h)}.}:
\begin{equation}
P(h) \rightarrow L_{n}(l_{n}) ... L_1(l_1) H(h) R_1(r_1) ... R_{m}(r_{m})
\label{eq-rule}
\end{equation}
$H$ is the head-child of the phrase, which inherits the
head-word $h$ from its parent $P$. $L_1 ... L_n$ and $R_1 ... R_m$ are
left and right modifiers of $H$. Either $n$ or $m$ may be zero,
and $n=m=0$ for unary rules. Figure~\ref{fig-lextree} 
shows a tree which will be used as an example throughout this paper.

\begin{figure}[htb]
\centerline{\psfig{figure=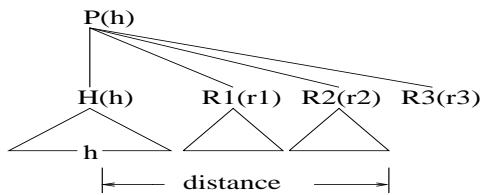,height=1in,width=2.5in}}
\caption{The next child, $R_3(r_3)$, is generated with probability
${\cal P}(R_3(r_3) \:  | \:  P,H,h,distance_r(2))$. The $distance$ is a function of the
surface string from the word after $h$ to the last word of $R_2$,
inclusive. In principle the model could condition on any structure
dominated by $H$, $R_1$ or $R_2$.} 
\label{fig-distance}
\end{figure}

The addition of lexical heads leads to an enormous number of potential rules, 
making direct estimation of ${\cal P}(RHS\:  | \: LHS)$ infeasible 
because of sparse data problems. We decompose
the generation of the {\tt RHS} of a rule such as~(\ref{eq-rule}),
given the {\tt LHS}, into
three steps --- first generating the head, then 
making the independence assumptions that the left and
right modifiers are generated by separate $0^{th}$-order markov
processes\footnote{An exception is the first rule in the tree,
{\tt TOP $\rightarrow$ H(h)}, which has probability $P_{TOP}(H,h|TOP)$}:

\begin{enumerate}
\item Generate the head constituent label 
of the phrase, with probability ${\cal P}_H(H\:  | \: P,h)$.
\item Generate modifiers to the right of the head with probability
$\prod_{i=1..m+1} {\cal P}_R(R_i(r_i) \:  | \:  P,h,H)$. $R_{m+1}(r_{m+1})$ is defined as
$STOP$ --- the $STOP$ symbol is added to the vocabulary of
non-terminals, and the model stops generating right modifiers when it
is generated.
\item Generate modifiers to the left of the head 
%in the same way,
with probability $\prod_{i=1..n+1}{\cal P}_L(L_i(l_i)\:  | \: P,h,H)$, where 
$L_{n+1}(l_{n+1}) = STOP$.
\end{enumerate}
For example, the probability of the rule
{\tt S(bought) -> N{\cal P}(week) N{\cal P}(Marks) V{\cal P}(bought)}
would be estimated as
\begin{tt}
\begin{eqnarray}
&&\hspace{-0.7cm}{\cal P}_h(\mbox{VP} \:  | \:  \mbox{S,bought}) \times
{\cal P}_l(\mbox{N{\cal P}(Marks)} \:  | \:  \mbox{S,VP,bought}) \times \nonumber \\
&&\hspace{-0.7cm}{\cal P}_l(\mbox{N{\cal P}(week)} \:  | \:  \mbox{S,VP,bought}) \times 
{\cal P}_l(\mbox{STOP} \:  | \:  \mbox{S,VP,bought}) \times \nonumber \\
&&\hspace{-0.7cm}{\cal P}_r(\mbox{STOP} \:  | \:  \mbox{S,VP,bought}) \nonumber 
\end{eqnarray}
\end{tt}
We have made the $0^{th}$ order markov assumptions
\begin{eqnarray}
{\cal P}_l(L_i(l_i) \:  | \:  H,P,h,L_1(l_1)...L_{i-1}(l_{i-1})) = \nonumber \\
{\cal P}_l(L_i(l_i) \:  | \:  H,P,h) \\
{\cal P}_r(R_i(r_i) \:  | \:  H,P,h,R_1(r_1)...R_{i-1}(r_{i-1})) = \nonumber \\
{\cal P}_r(R_i(r_i) \:  | \:  H,P,h) 
\end{eqnarray}
but in general the probabilities could be conditioned on any of the preceding
modifiers. In fact, if the derivation order is fixed to be depth-first ---
that is, 
each modifier recursively generates the sub-tree below it before the next
modifier is generated --- then the model can also condition on any structure
{\em below} the preceding modifiers. For the moment we
exploit this by making the approximations
\begin{eqnarray}
{\cal P}_l(L_i(l_i) \:  | \:  H,P,h,L_1(l_1)...L_{i-1}(l_{i-1})) = \nonumber \\
 {\cal P}_l(L_i(l_i) \:  | \:  H,P,h,distance_l(i-1))\\
{\cal P}_r(R_i(r_i) \:  | \:  H,P,h,R_1(r_1)...R_{i-1}(r_{i-1})) = \nonumber \\
{\cal P}_r(R_i(r_i) \:  | \:  H,P,h,distance_r(i-1))
\end{eqnarray}
where $distance_l$ and $distance_r$ 
are functions of the surface string from the head word to the edge of
the constituent (see figure~\ref{fig-distance}). The distance measure
is the same as in \cite{collins}, a vector with the 
following 3 elements: (1) is the string of zero length? (Allowing the
model to learn a preference for right-branching structures);
(2) does the string contain a verb? (Allowing the model to learn a
preference for modification of the most recent verb).
(3) Does the string contain 0, 1, 2 or $>2$
commas? (where a comma is anything tagged as ``,'' or ``:''). 

%\subsection{Predicate Argument Structure}

%Most natural language applications which use the output from a parser will
%need to extract predicate-argument structure. However, previous work
%on treebank parsing (collins,spatter,charniak) has ignored useful
%information in the treebank annotations: the
%complement/adjunct distinction, and information about wh-movement.
%In this section we show how this information can be integrated
%into the parsing model.

\subsection{Model 2: The complement/adjunct distinction and subcategorisation}

\begin{figure}[htb]
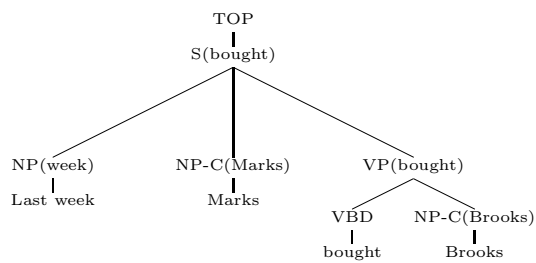

\begin{tiny}
\begin{center}
\vspace{-1ex}
\Tree  [.TOP [.S(bought) [.NP(week)  $\mbox{Last week}$ ]
[.NP-C(Marks)  Marks  ]
[.VP(bought)  [.VBD bought ] [.NP-C(Brooks) Brooks ] ] ] ]
\end{center}
\end{tiny}
\caption{A tree with the ``-C'' suffix used to identify complements.
``Marks'' and ``Brooks'' are in subject and object position
respectively. ``Last week'' is an adjunct.}
\vspace{-1ex}
\label{fig-comp}
\end{figure}

The tree in figure~\ref{fig-lextree} is an example of the importance of the
complement/adjunct distinction.
It would be useful to identify ``Marks'' as a subject, and ``Last week''
as an adjunct (temporal modifier), but this distinction is not made in
the tree, as both NPs are in the same position\footnote{Except ``Marks''
is closer to the VP, but note that ``Marks'' is also the subject in
``Marks last week bought Brooks''.} (sisters to a
VP under an S node). From here on we will identify complements by
attaching a ``-C'' suffix to non-terminals --- figure~\ref{fig-comp}
gives an example tree.

A post-processing stage could add this detail to the parser output,
but we give two reasons for making the distinction while parsing:
 First, identifying complements
is complex enough to warrant a probabilistic treatment. Lexical information is
needed --- for example, knowledge that ``week'' is likely to be a
temporal modifier. Knowledge about subcategorisation preferences ---
for example that a verb takes exactly one subject --- is also required.
These problems are not restricted to NPs, compare ``The spokeswoman said
(SBAR that the asbestos was dangerous)'' vs. ``Bonds beat short-term
investments (SBAR because the market is down)'', where an SBAR headed
by ``that'' is a complement, but an SBAR headed by ``because'' is an adjunct.

The second reason for making the
complement/adjunct distinction while parsing is that it may help
parsing accuracy. The assumption that complements
are generated independently of each other often leads to incorrect
parses --- see figure~\ref{fig-ind} for further explanation. 

\begin{figure*}[htb]
\begin{enumerate}
\begin{tiny}
\item (a) Incorrect \hspace{-1in} \Tree [.S [.NP-C Dreyfus ] [.NP-C $\mbox{the best fund}$ ] [.VP was
[.ADJP low ] ] ]
\hspace{-0.06in} (b) Correct \hspace{-0.5in} \Tree [.S [.NP-C [.NP Dreyfus ] [.NP $\mbox{the best fund}$ ] ] [.VP was
[.ADJP low ] ] ]
\item (a) Incorrect \hspace{-1in} \Tree [.S [.NP-C $\mbox{The issue}$ ] [.VP was [.NP-C $\mbox{a bill}$ ] [.VP-C
funding [.NP-C Congress ] ] ] ]
\hspace{-0.5in} (b) Correct \hspace{-0.5in} \Tree [.S [.NP-C $\mbox{The issue}$ ] [.VP was [.NP-C [.NP $\mbox{a bill}$ ] [.VP
funding [.NP-C Congress ] ] ] ] ]
\end{tiny}
\end{enumerate}
\vspace{-3ex}
\caption{Two examples where the assumption that
modifiers are generated independently of each other
leads to errors.
In (1) the probability of generating both ``Dreyfus'' and ``fund'' as
subjects, {\tt ${\cal P}(\mbox{NP-C(Dreyfus)} \:  | \:  \mbox{S,VP,was})
 * {\cal P}(\mbox{NP-C(fund)} \:
| \:  \mbox{S,VP,was})$} is unreasonably high.
(2) is similar: {\tt ${\cal P}(\mbox{NP-C(bill),VP-C(funding)} \:  | \:  
\mbox{VP,VB,was})
= {\cal P}(\mbox{NP-C(bill)} \:  | \:  \mbox{VP,VB,was}) *
{\cal P}(\mbox{VP-C(funding)} \:  | \:  \mbox{VP,VB,was})$} is a bad independence assumption.
}
\label{fig-ind}
\end{figure*}

\subsubsection{Identifying Complements and Adjuncts in the Penn Treebank}

%The semantic tags on the WSJ treebank (ref) lead to a definition of
%complement/adjunct distinction. 
We add the ``-C'' suffix to all
non-terminals in training data which satisfy the following conditions:

\begin{enumerate}

\item The non-terminal
must be: (1) an NP, SBAR, or S whose parent is an S; (2) an NP,
SBAR, S, or VP whose parent is a VP; or (3) an S whose parent is an SBAR.

\item The non-terminal must {\em not} have one of the following semantic tags:
ADV, VOC, BNF, DIR, EXT, LOC, MNR, TMP, CLR or PRP. See \cite{marcus2} for an
explanation of what these tags signify. For example,
the NP ``Last
week'' in figure~\ref{fig-lextree} 
would have the TMP (temporal) tag; and the SBAR in 
``(SBAR because the market is down)'', would have the ADV (adverbial)
tag.

\end{enumerate}

In addition, the first child following the head of a
prepositional phrase is marked as a complement. 

\subsubsection{Probabilities over Subcategorisation Frames}

The model could be retrained on training data with the enhanced set of
non-terminals, and it
might learn the lexical properties which distinguish
complements and adjuncts (``Marks'' vs ``week'', or ``that''
vs. ``because''). However, it would still suffer from the bad independence
assumptions illustrated in figure~\ref{fig-ind}.
To solve these kinds of problems, the generative process is extended
to include a probabilistic choice of left and right subcategorisation frames:

\begin{enumerate}

\item Choose a head $H$ with probability ${\cal P}_H(H\:  | \: P,h)$.

\item Choose left and right subcat frames, $LC$ and $RC$, with
probabilities ${\cal P}_{lc}(LC\:  | \: P,H,h)$ and ${\cal P}_{rc}(RC\:  | \: P,H,h)$. Each subcat frame
is a multiset\footnote{A multiset, or bag, is a set which may contain
duplicate non-terminal labels.}
specifying the complements which the head requires
in its left or right modifiers. 

\item Generate the left and right modifiers with
probabilities ${\cal P}_l(L_i,l_i \:  | \:  H,P,h,distance_l(i-1),LC)$ and 
${\cal P}_r(R_i,r_i \:  | \:  H,P,h,distance_r(i-1),RC)$ respectively. 
Thus the subcat
requirements are added to the conditioning context. As
complements are generated they are removed from the appropriate subcat
multiset. Most importantly, the probability of generating the $STOP$
symbol will be 0 when the subcat frame is {\em non-empty}, and the
probability of generating a complement will be 0 when it is not in the 
subcat frame; thus all and only
the required complements will be generated.
\end{enumerate}
The probability of the phrase 
{\tt S(bought) -> N{\cal P}(week) NP-C(Marks) V{\cal P}(bought)}
is now:
{\tt \begin{eqnarray*}
&&\hspace{-0.7cm}{\cal P}_h(\mbox{VP} \:  | \:  \mbox{S,bought}) \times \\
&&\hspace{-0.7cm}{\cal P}_{lc}(\{\mbox{NP-C}\} \:  | \:  \mbox{S,VP,bought}) \times 
{\cal P}_{rc}(\{\} \:  | \:  \mbox{S,VP,bought})  \times \\
&&\hspace{-0.7cm}{\cal P}_l(\mbox{NP-C(Marks)} \:  | \:  \mbox{S,VP,bought},\{\mbox{NP-C}\}) \times \\
&&\hspace{-0.7cm}{\cal P}_l(\mbox{N{\cal P}(week)} \:  | \:  \mbox{S,VP,bought},\{\}) \times \\
&&\hspace{-0.7cm}{\cal P}_l(\mbox{STOP} \:  | \:  \mbox{S,VP,bought},\{\}) \times \\
&&\hspace{-0.7cm}{\cal P}_r(\mbox{STOP} \:  | \:  \mbox{S,VP,bought},\{\}) 
\end{eqnarray*} }
Here the head initially decides to take a single {\tt NP-C} (subject) 
to its left, and no complements to its right. {\tt NP-C(Marks)} is
immediately generated as the required subject, and {\tt NP-C} is removed
from $LC$, leaving it empty when the next modifier, {\tt N{\cal P}(week)} is
generated. 
The incorrect structures in figure~\ref{fig-ind} should now have low
probability because {\tt ${\cal P}_{lc}(\{\mbox{NP-C,NP-C}\} \:  | \: 
\mbox{S,VP,bought})$ }and {\tt ${\cal P}_{rc}(\{\mbox{NP-C,VP-C}\}\:  | \: \mbox{VP,VB,was})$} are small.
\begin{figure*}[tb]
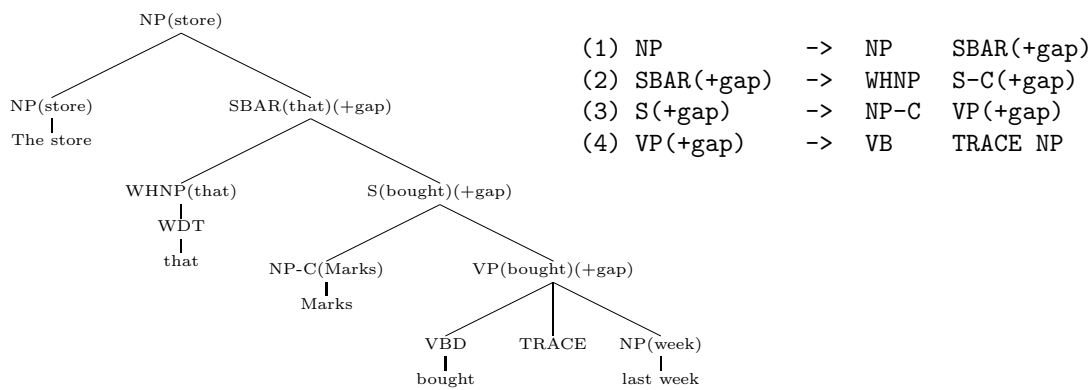

\begin{tiny}
\vspace{-0.5in}
\hspace{-0.5in}\Tree  [.NP(store)  [.NP(store)  $\mbox{The store}$
] [.SBAR(that)(+gap)  [.WHNP(that)  [.WDT that ]  ]  [.S(bought)(+gap)
[.NP-C(Marks)  Marks ]
[.VP(bought)(+gap)  [.VBD bought ]  TRACE  [.NP(week)
$\mbox{last week}$
] ] ] ] ]
\end{tiny}
\begin{tt}
\hspace{-1.5in}
\begin{tabular}{lllll}
&&&&\\
&&&&\\
&&&&\\
&&&&\\
&&&&\\
(1) NP         &->&   NP  &   SBAR(+gap) &\\
(2) SBAR(+gap) &->&   WHNP &  S-C(+gap) &\\
(3) S(+gap)    &->&   NP-C &  VP(+gap) &\\
(4) VP(+gap)   &->&   VB   &  TRACE     NP \\
\end{tabular}
\end{tt}
\caption{A $+gap$ feature can be added to non-terminals to describe NP
extraction.
The top-level NP initially generates an SBAR modifier, but specifies that it
must contain an NP trace by adding the {\em +gap} feature. The gap is
then passed down through the tree, until it is discharged as a $TRACE$
complement to the right of $bought$. 
}
\label{fig-gap}
\end{figure*}

\subsection{Model 3: Traces and Wh-Movement}

Another obstacle to extracting predicate-argument structure from
parse trees is wh-movement. This section describes a
probabilistic treatment of extraction from relative clauses.
Noun phrases are most often extracted from subject position, object
position, or from within PPs:
\newtheorem{sentence}{Example}
\begin{sentence}
The store (SBAR which TRACE bought Brooks Brothers)
\end{sentence}
\begin{sentence}
The store (SBAR which Marks bought TRACE)
\end{sentence}
\begin{sentence}
The store (SBAR which Marks bought Brooks Brothers from TRACE)
\end{sentence}
It might be possible to write rule-based patterns
which identify traces in a parse tree. However, we argue again that this
task is best integrated into the parser: the task is complex enough to
warrant a probabilistic treatment, and integration may help parsing
accuracy. A couple of complexities are that modification by an SBAR
does not always involve extraction (e.g., ``the fact (SBAR that besoboru is
played with a ball and a bat)''), and 
it is not uncommon for extraction to occur through several
constituents, (e.g.,
``The changes (SBAR that he said the government was
prepared to make TRACE)'').

The second reason for an integrated treatment of traces is to improve the
parameterisation of the model. In particular, the subcategorisation 
probabilities are smeared by extraction. In examples 1, 2 and 3
above `bought' is a transitive verb, but without knowledge of traces
example 2 in training data will contribute to the probability
of `bought' being an intransitive verb.

Formalisms similar to GPSG \cite{gpsg} handle NP extraction by
adding a {\em gap} feature to each non-terminal in the tree, and
propagating gaps through the tree until they are finally discharged as
a trace complement (see figure~\ref{fig-gap}).
In extraction cases the Penn treebank annotation co-indexes
a TRACE with the WHNP head of the SBAR, so it is straightforward to
add this information to trees in training data.

Given that
the LHS of the rule has a gap, there are 3 ways that the gap can be 
passed down to the RHS:

\begin{description}

\item[Head] The gap is passed to the head of the phrase, as in rule
(3) in figure~\ref{fig-gap}.

\item[Left, Right] The gap is passed on recursively to one of the left or right
modifiers of the head, or is discharged as a $trace$ argument to the
left/right of the head. In rule (2) it is passed on to a right
modifier, the S complement. In rule (4) 
a $trace$ is generated to the right of the head VB.

\begin{figure*}[tb]
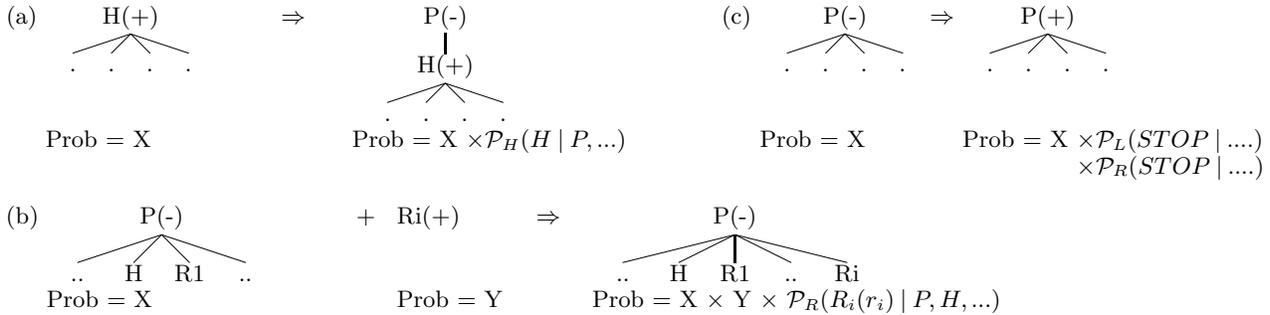

\begin{small}
\begin{tabbing}
(a) \= \Tree [.H(+) . . . .  ] \hspace{-0.4cm} $\Rightarrow$ \hspace{0.4cm} \= \Tree [.P(-) [.H(+) . . . .  ] ]  
\hspace{0.5in} (c) \= \Tree [.P(-) . . . . ]  \hspace{-0.5in} $\Rightarrow$ \=
\Tree[.P(+) . . . . ] \\
\> Prob = X \> Prob = X $\times {\cal P}_H(H\:  | \: P,...)$ \> Prob = X \> Prob = X
\= $\times {\cal P}_L(STOP\:  | \: ....) $ \\
\> \> \> \> \> $\times {\cal P}_R(STOP\:  | \: ....)$
\end{tabbing}
\begin{tabbing}
(b) \= \Tree [.P(-) .. H R1 .. ] \hspace{-1.5ex} +  \hspace{0.5ex} \= Ri(+) \hspace{0.8cm} $\Rightarrow$
\hspace{0.2cm} \= \Tree [.P(-) .. H R1 .. Ri ] \\
\>Prob = X \> Prob = Y \> Prob = X $\times$ Y $\times$
${\cal P}_{R}(R_i(r_i)\:  | \: P,H,...)$
\end{tabbing}
\end{small}
\vspace{-3ex}
\caption{The life of a constituent in the chart.
$(+)$ means a constituent is
complete (i.e. it includes the stop probabilities), $(-)$ means a constituent
is incomplete. (a) a new constituent is started
by projecting a complete rule upwards; (b)
the constituent then takes left and right modifiers (or none if it is
unary). (c) finally, $STOP$ probabilities are added to complete the
constituent. }
\label{fig-parse}
\end{figure*}

\end{description}
\begin{table*}[tb]
\begin{small}
\begin{center}
\begin{tabular}{|c||c|c|c|c|}
\hline
Back-off&${\cal P}_H(H\:  | \: ...)$ & ${\cal P}_G(G\:  | \: ...)$ & 
${\cal P}_{L1}(L_i(lt_i)\:  | \: ...)$ &
${\cal P}_{L2}(lw_i\:  | \: ...)$ \\
Level&&${\cal P}_{LC}(LC\:  | \: ...)$&
${\cal P}_{R1}(R_i(rt_i)\:  | \: ...)$ & 
${\cal P}_{R2}(rw_i\:  | \: ...)$ \\
&&${\cal P}_{RC}(RC\:  | \: ...)$ &&\\
\hline
\hline
1 & P, w, t & P, H, w, t & P, H, w, t, $\Delta$, LC &
$L_i$, $lt_i$, P, H, w, t, $\Delta$, LC \\
2 & P, t & P, H, t & P, H, t, $\Delta$, LC &
$L_i$, $lt_i$, P, H, t, $\Delta$, LC \\
3 & P & P, H & P, H, $\Delta$, LC &
$L_i$, $lt_i$ \\
4&---&---&---&
$lt_i$ \\
\hline
\end{tabular}
\end{center}
\end{small}
\caption{The conditioning variables for each level of back-off. For
example,  ${\cal P}_H$ estimation interpolates $e_1 = {\cal P}_H(H\:  | \: P,w,t)$,
$e_2 = {\cal P}_H(H\:  | \: P,t)$, and $e_3 = {\cal P}_H(H\:  | \: P)$.
$\Delta$ is the distance measure.}
\label{tab-backoff}
\end{table*}

We specify a parameter ${\cal P}_G(G\:  | \: P,h,H)$ where $G$ is either {\bf
Head}, {\bf Left} or {\bf Right}.
The generative process is extended to choose between these cases after
generating the head of the phrase. The rest of the phrase is then generated
in different ways depending on how the gap is propagated:
In the {\bf Head} case the left and right modifiers
are generated as normal. 
In the {\bf Left, Right} cases
a {\em gap} requirement is added to either the left or right
SUBCAT variable. This requirement is fulfilled (and removed from
the subcat list) when a trace or a modifier non-terminal which
has the {\em +gap} feature is generated. 
%Below are a couple of examples of
%gap-propagation: 
For example, 
Rule (2), {\tt SBAR(that)(+gap) -> WHNP(that) S-C(bought)(+gap)}, has
probability
{\tt \begin{eqnarray*}
&&\hspace{-0.7cm}{\cal P}_h(\mbox{WHNP} \:  | \: \mbox{SBAR,that}) \times {\cal P}_G(\mbox{Right} \:  | \: \mbox{SBAR,WHNP,that})
\times \\
&&\hspace{-0.7cm}{\cal P}_{LC}(\{\} \:  | \:  \mbox{SBAR,WHNP,that}) \times \\
&&\hspace{-0.7cm}{\cal P}_{RC}(\{\mbox{S-C}\} \:  | \:  \mbox{SBAR,WHNP,that})
\times \\
&&\hspace{-0.7cm}{\cal P}_R(\mbox{S-C(bought)(+gap)} \:  | \:  \mbox{SBAR,WHNP,that}, \{\mbox{S-C,+gap}\}) \times \\
&&\hspace{-0.7cm}{\cal P}_R(\mbox{STOP} \:  | \: \mbox{SBAR,WHNP,that},\{\}) \times \\
&&\hspace{-0.7cm}{\cal P}_L(\mbox{STOP} \:  | \:  \mbox{SBAR,WHNP,that},\{\}) 
\end{eqnarray*}}
Rule (4), {\tt VP(bought)(+gap) -> VB(bought) TRACE NP(week)}, has
probability 
{\tt \begin{eqnarray*}
&&\hspace{-0.7cm}{\cal P}_h(\mbox{VB} \:  | \: \mbox{VP,bought}) \times {\cal P}_G(\mbox{Right} \:  | \: \mbox{VP,bought,VB}) \times \\
&&\hspace{-0.7cm}{\cal P}_{LC}(\{\} \:  | \:  \mbox{VP,bought,VB}) \times
{\cal P}_{RC}(\{\mbox{NP-C}\} \:  | \:  \mbox{VP,bought,VB}) 
\times \\
&&\hspace{-0.7cm}{\cal P}_R(\mbox{TRACE} \:  | \:  \mbox{VP,bought,VB}, \{\mbox{NP-C, +gap}\}) \times \\
&&\hspace{-0.7cm}{\cal P}_R(\mbox{N{\cal P}(week)} \:  | \:  \mbox{VP,bought,VB}, \{\}) \times \\
&& \hspace{-0.7cm}{\cal P}_L(\mbox{STOP} \:  | \:  \mbox{VP,bought,VB},\{\}) \times \\
&& \hspace{-0.7cm}{\cal P}_R(\mbox{STOP} \:  | \:  \mbox{VP,bought,VB},\{\})
\end{eqnarray*} }
In rule (2) {\tt Right} is chosen, so the $+gap$ requirement
is added to $RC$. Generation of {\tt S-C(bought)(+gap) } fulfills both
the {\tt S-C} and $+gap$ requirements in $RC$. In rule (4) {\tt Right}
is chosen again. Note that generation of $trace$ satisfies both the
{\tt NP-C} and $+gap$ subcat requirements.

\section{Practical Issues}
\label{sec-prac}

\subsection{Smoothing and Unknown Words}

Table~\ref{tab-backoff}
shows the various levels of back-off for each type of parameter 
in the model. Note that we decompose
${\cal P}_{L}(L_i(lw_i,lt_i)\:  | \: P,H,w,t,\Delta,LC)$ (where $lw_i$ and
$lt_i$ are the word and POS tag generated with non-terminal $L_i$,
$\Delta$ is the distance measure)
into the product ${\cal P}_{L1}(L_i(lt_i)\:  | \: P,H,w,t,\Delta,LC) \times
{\cal P}_{L2}(lw_i\:  | \: L_i,lt_i,P,H,w,t,\Delta,LC)$, and then smooth these two
probabilities separately (Jason Eisner, p.c.). In each case\footnote{Except
cases $L_2$ and $R_2$, which have 4 levels, so that
$e = \lambda_1 e_1 + (1-\lambda_1) (\lambda_2 e_2 + (1-\lambda_2) 
( \lambda_3 e_3 + (1 - \lambda_3) e_4)
)$.}
the final estimate is 

\[e = \lambda_1 e_1 + (1-\lambda_1) (\lambda_2 e_2 +
(1-\lambda_2) e_3)\]
where $e_1$, $e_2$ and $e_3$ are maximum likelihood estimates with the
context at levels 1, 2 and 3 in the table, and $\lambda_1$,
$\lambda_2$ and $\lambda_3$ are smoothing parameters where $0 \leq
\lambda_i \leq 1$. All words occurring  
less than 5 times in training data, and words in
test data which have never been seen in training, are replaced
with the ``UNKNOWN'' token. This allows the model to robustly handle the
statistics for rare or new words.

%The $\lambda$ values are
%chosen to be $\frac{f}{5+f}$ for ${\cal P}_H$, ${\cal P}_G$, ${\cal P}_{LC}$ and ${\cal P}_{RC}$,
%and $\frac{f}{5u+f}$ for ${\cal P}_{L1}$, ${\cal P}_{R1}$, ${\cal P}_{L2}$ and 
%${\cal P}_{R2}$, where $f$ = the denominator of the more specific estimate,
%and $u$ is the number of unique outcomes seen in the more specific
%distribution.

%Level&${\cal P}_H(H\:  | \: P,w,t)$ & ${\cal P}_G(G\:  | \: P,H,w,t)$ & 
%${\cal P}_{L1}(L_i(lt_i)\:  | \: P,H,w,t,\Delta,LC)$ &
%${\cal P}_{L2}(lw_i\:  | \: L_i,lt_i,P,H,w,t,\Delta,LC)$ \\
%&&${\cal P}_{LC}(LC\:  | \: P,H,w,t)$&
%${\cal P}_{R1}(R_i(rt_i)\:  | \: P,H,w,t,\Delta,RC)$ & 
%${\cal P}_{R2}(rw_i\:  | \: R_i,rt_i,P,H,w,t,\Delta,RC)$ \\
%&&${\cal P}_{RC}(RC\:  | \: P,H,w,t)$ &&\\

\subsection{Part of Speech Tagging and Parsing}

Part of speech tags are generated along with the words in this
model. When parsing, the POS tags allowed for each word are
limited to those which have been seen in training data for that
word. For unknown words, the output from the tagger described in
\cite{adwait} is used as the single possible tag for that word.
A CKY style dynamic programming chart parser is used to find the
maximum probability tree for each sentence (see figure~\ref{fig-parse}).

\begin{table*}[t]
\begin{center}
\begin{tabular}{|c||c|c|c|c|c||c|c|c|c|c|}
\hline
MODEL&\multicolumn{5}{|c||}{$\leq$ 40 Words (2245 sentences)}&\multicolumn{5}{|c|}{$\leq$ 100 Words (2416 sentences)}\\ \cline{2-11}
&LR&LP&CBs&$0$ CBs&$\leq 2$ CBs&LR&LP&CBs&$0$ CBs&$\leq 2$ CBs\\
\hline
\hline
\cite{magerman}&84.6\%&84.9\%&1.26&56.6\%&81.4\%&84.0\%&84.3\%&1.46&54.0\%&78.8\%\\
\cite{collins}&85.8\%&86.3\%&1.14&59.9\%&83.6\%&85.3\%&85.7\%&1.32&57.2\%&80.8\%\\
\hline
Model 1&87.4\%&88.1\%&0.96&65.7\%&86.3\%&86.8\%&87.6\%&1.11&63.1\%&84.1\%\\
Model 2&88.1\%&88.6\%&0.91&66.5\%&86.9\%&87.5\%&88.1\%&1.07&63.9\%&84.6\%\\
Model 3&88.1\%&88.6\%&0.91&66.4\%&86.9\%&87.5\%&88.1\%&1.07&63.9\%&84.6\%\\
\hline
\hline
\end{tabular}
\caption{Results on Section 23 of the WSJ Treebank.
{\bf LR/LP} = labeled
recall/precision. {\bf CBs} is the average number of crossing brackets per
sentence. {\bf 0 CBs, $\leq 2$ CBs} are the percentage of sentences with 0
or $\leq 2$ crossing brackets respectively.
}
\vspace{-4ex}
\end{center}
\label{tab-results}
\end{table*}

\section{Results}
\label{sec-results}

The parser was trained on sections~02~-~21 of the Wall Street Journal 
portion of the Penn Treebank \cite{marcus}
(approximately 40,000 sentences), and tested on 
section~23 (2,416 sentences). 
We use the PARSEVAL measures 
\cite{black} to compare performance:

\begin{description}

\item[Labeled Precision =]\begin{Large}
$\frac{number \:  of \:  correct \:  constituents \:  in \:  proposed \:  parse}
{number \:  of \:  constituents \:  in \:  proposed \:  parse}$ \end{Large}

\item[Labeled Recall =]\begin{Large}
$\frac{number \:   of \:   correct \:   constituents \:   in \:   proposed \:   parse}
{number \:   of \:   constituents \:   in \:   treebank \:   parse}$ \end{Large}

\item[Crossing Brackets =] number of constituents which violate constituent 
boundaries with a constituent in the treebank parse.

\end{description}

For a constituent to be `correct' it must span the same set of 
words (ignoring punctuation, i.e. all tokens tagged as commas, colons
or quotes) and have the same label\footnote{\cite{magerman} collapses {\tt ADVP} and
{\tt PRT} to the same label, for comparison
we also removed this distinction when calculating scores.} 
as a constituent in the treebank parse. Table~2 shows
the results for Models 1, 2 and 3. 
The precision/recall of the traces found by Model 3 was 93.3\%/90.1\%
(out of 436 cases in section 23 of the treebank),
where three criteria must be met for a trace to be ``correct'': (1) it
must be an argument to the correct head-word; (2) it must be in the
correct position in relation to that head word (preceding or
following); (3) it must be dominated by the correct non-terminal
label. For example, in figure~\ref{fig-gap} the trace is an argument
to {\bf bought}, which it {\bf follows}, and it is dominated by a {\bf VP}.
Of the 436 cases, 342 were string-vacuous extraction from subject position,
recovered with 97.1\%/98.2\% precision/recall; and 94 were longer distance 
cases, recovered with 76\%/60.6\% precision/recall
\footnote{We exclude infinitival relative clauses from these figures, 
for example ``I called a plumber TRACE to fix the sink'' where `plumber' 
is co-indexed with the trace subject of the infinitival. The 
algorithm scored 41\%/18\% precision/recall on the 60 cases in section 23
--- but infinitival relatives are extremely difficult even for
human annotators to distinguish from purpose clauses (in this case, the
infinitival could be a purpose clause modifying `called') (Ann Taylor, p.c.)}.

\subsection{Comparison to previous work}
\label{sec-compare}

Model 1 is similar in structure to \cite{collins}
---  the major differences being that the ``score'' for each bigram
dependency is ${\cal P}_l(L_i,l_i | H,P,h,distance_l)$ 
rather than 
${\cal P}_l(L_i,P,H \:  | \:  l_i,h,distance_l)$, and that 
there are the additional probabilities of generating the head and
the $STOP$ symbols for each constituent.
However, Model 1 has some advantages which may account for the improved
performance. The model in \cite{collins} is deficient, that is for most
sentences $S$, $\sum_T {\cal P}(T\:  | \: S) < 1$, 
because probability mass is lost to dependency structures
which violate the hard constraint that no links may cross. 
%This
%mathematical problem may lead to a drop in performance. 
For reasons we
do not have space to describe here, Model 1 has advantages in its
treatment of unary rules and the distance measure. The generative model
can condition on any structure that has been previously
generated --- we exploit this in models 2 and 3 ---  
whereas \cite{collins} is restricted to conditioning on features of
the surface string alone.

\cite{charniak} also uses a lexicalised generative model. In our notation,
he decomposes ${\cal P}(RHS_i\:  | \: LHS_i)$ as ${\cal P}(R_n ...R_1 H L_1 .. L_m \:  | \:  P,h)
\times \prod_{i=1..n} {\cal P}(r_i\:  | \: P,R_i,h) \times \prod_{i=1..m}
{\cal P}(l_i\:  | \: P,L_i,h)$. 
The Penn treebank annotation style leads to 
a very large number of context-free rules, so that
directly estimating ${\cal P}(R_n ...R_1 H L_1
.. L_m \:  | \:  P,h)$ may lead to sparse data problems, or problems
with coverage (a rule which has never been seen in training
may be required for a test data sentence).
The complement/adjunct distinction and traces increase the number of
rules, compounding this problem.

\cite{eisner} proposes 3 dependency models, and gives results that
show that a generative model similar to Model 1 performs best of the three.
However, a pure dependency model omits non-terminal information, which
is important. For example, ``hope'' is likely to generate a
{\tt VP(TO)} modifier (e.g., 
I hope [VP to sleep]) whereas ``require'' is likely
to generate an {\tt S(TO)} modifier (e.g., I require [S Jim to
sleep]), but
omitting non-terminals conflates these two cases, giving high probability to
incorrect structures such as ``I hope [Jim to sleep]'' or
``I require [to sleep]''.
\cite{alshawi} extends a generative dependency
model to include an additional state variable which is equivalent
to having non-terminals --- his suggestions may be close to our models 1 and
2, but he does not fully specify the details of his model, and doesn't give
results for parsing accuracy. \cite{miller} describe a model
where the {\tt RHS} of a rule is generated by a Markov process, although
the process is not head-centered. They increase the set of
non-terminals by adding semantic labels rather than by adding lexical
head-words. 

\cite{magerman,ibm} describe a history-based approach which uses
decision trees to estimate ${\cal P}(T|S)$.
Our models use much less sophisticated n-gram estimation methods, and
might well benefit from methods such as decision-tree estimation which
could condition on richer history than just surface distance.

There has recently been interest in using dependency-based parsing
models in speech recognition, for example \cite{stolcke}.
It is interesting to note that Models 1, 2 or 3 could be used as language
models. The probability for
any sentence can be estimated as ${\cal P}(S) = \sum_T {\cal P}(T,S)$,
or (making a Viterbi approximation for efficiency reasons) as
${\cal P}(S) \approx {\cal P}(T_{best},S)$. We intend to perform experiments
to compare the perplexity of the various models, and a structurally similar
`pure' PCFG\footnote{Thanks to one of the anonymous reviewers for suggesting 
these experiments.}.

\section{Conclusions}

\label{sec-conc}

This paper has proposed a generative, lexicalised, probabilistic
parsing model. We have shown that linguistically fundamental ideas,
namely subcategorisation and wh-movement, can be given a 
statistical interpretation. This improves parsing
performance, and, more importantly, adds useful information
to the parser's output.

\vspace{-2ex}
\section{Acknowledgements}

I would like to thank Mitch Marcus, Jason Eisner, Dan Melamed and Adwait
Ratnaparkhi for many useful discussions, and comments on earlier versions
of this paper. This work has also benefited greatly from suggestions and advice
from Scott Miller.

\end{document}